\documentclass[runningheads]{llncs}
\usepackage{adjustbox}
\usepackage{graphicx}
\usepackage{multirow}
\usepackage[figuresleft]{rotating}
\begin{document}
\title{Comparison of Outlier Detection Techniques for Structured Data}
\author{Amulya Agarwal\inst{1} \and
Nitin Gupta \inst{2}}
\authorrunning{Agarwal et al.}

\institute{Indian Institutes of Information Technology Delhi, India \\
\email{amulya20004@iiitd.ac.in}\\
\and
IBM Research AI, India\\
\email{ngupta47@in.ibm.com}}

\maketitle              % typeset the header of the contribution
\begin{abstract}
An outlier is an observation or a data point that is far from rest of the data points in a given dataset or we can be said that an outlier is away from the center of mass of observations. Presence of outliers can skew statistical measures and data distributions which can lead to misleading representation of the underlying data and relationships. It is seen that the removal of outliers from the training dataset before modeling can give better predictions. With the advancement of machine learning, the outlier detection models are also advancing at a good pace. The goal of this work is to highlight and compare some of the existing outlier detection techniques for the data scientists to use that information for outlier algorithm selection while building a machine learning model.

\end{abstract}
\section{Introduction}
% An outlier is an observation that deviates so much from other observations as to arouse suspicion that it was generated by a different mechanism. The presence of outliers in data degrades the model performance \cite{outlier_impact}. It is therefore important to identify and remove anomalous data objects from the general data distribution in order to preserve the integrity of data science models. Outlier detection can help in several industry applications like fraud detection in finance \cite{ahmed2016survey}, fault diagnosis in mechanics \cite{shin2005one}, intrusion detection in network security \cite{garcia2009anomaly} and pathology detection in medical imaging \cite{baur2018deep}. Some other applications are discovery of criminal activities in electronic commerce, weather prediction, marketing and customer segmentation etc.

% \subsection{What is the outlier}
An outlier \cite{outlier_definition} is an observation that deviates so much from other observations as to arouse suspicion that it was generated by a different mechanism. Hence, the outliers skew the statistical measures and give a wrong representation of data and other relationships \cite{cousineau2010outliers}. This leads to poor fitting of classification and regression models on the dataset and hence poor predictions \cite{outlier_impact}. Outliers can be categorized \cite{zhao2019pyod} as: Univariate \cite{mowbray2019univariate} (in a distribution of values in a single feature space), multivariate \cite{leys2018detecting} (found in a n-dimensional space), point outliers \cite{kriegel2010outlier} (single data points that are far from the rest of the distribution), contextual outliers \cite{liang2016robust} (like the noise) or collective outliers (subsets of novelties). The main causes of generation of these outliers \cite{barnett1978study} are : Human errors like wrong entry of data, measuring instrument errors, errors while data extraction, experiment or data processing (like manipulation lead to far away points). Outlier may have been generated intentionally to test the detection methods or unintetionally by extracting or mixing data from wrong or various sources.

Detection of outlier is very important in itself as many applications require to determine whether the new data (observation) belong to the same data distribution or different. If it is from same distribution, is it called as outlier and if it is from the same distribution, it is called as an inlier \cite{hido2008inlier}. Detection of outlier is important as it can give additional information about localized anomalies in the whole system. Removing outliers from training data set before application of classification or regression models, may result in a better fit. Removal of outliers will help to assume the underlying data distribution in a better way and hence can result in better predictions on unseen test data. Removal of outliers finds its application in detecting fraudulent applications \cite{ahmed2016survey}, intrusion detection in network security \cite{garcia2009anomaly} and pathology detection in medical imaging \cite{baur2018deep}, monitoring the performance of computer networks, for example to detect network bottlenecks, monitoring processes to detect faults in motors, generators, pipelines or space instruments, identifying novel features or misclassified features, monitoring safety critical applications such as drilling or high-speed milling in time series, detecting unexpected entries and mislabelled data in a training data set \cite{article}. With the motivation of helping data scientist, we compare different outlier detection techniques in this work. This comparison can help the data scientist in outlier detection algorithm selection for building a better model.

The paper is organized as follows: Section \ref{outl} provides the brief summary of 15 different outlier detection techniques that we compare in this paper. Section \ref{res} discuss the dataset details, evaluation metrics, and experimental evaluation. Finally, we conclude in Section \ref{conc}.

\section{Outlier Detection Techniques}
\label{outl}
Several outlier detection algorithms have been proposed over the last decades, which includes (a) proximity based models \cite{5070231} - generally used when the locality of a data point is sparse, (b) linear models \cite{bounsiar2014one} - used for lower-dimensional subspace data as it utilizes inter attribute dependencies, (c) ensemble base models \cite{zhao2019lscp} - used to exploit the power of multiple outlier detection techniques through voting mechanism, and (d) probabilistic based models \cite{li2020copod} etc - used generally when multi-varaiate data is sparse. Table \ref{table1} provides categorization of different outlier detection algorithms that we plan to cover in this paper. In this section, we will provide details of various oultlier detection algorithms.

\begin{table}[t]
\begin{tabular}{ |c|c|c|c| } 
 \hline
\textbf{ Method} & \textbf{Category}\\ 
 \hline
 \hline
 Local Outlier Factor (LOF) & Proximity \\ 
\hline
 K Nearest Neighbors (KNN) & Proximity \\ 
\hline
Clustering-Based Local Outlier Factor (CBLOF) & Proximity  \\
\hline
Angle-Based Ouylier detection (ABOD)  & Proximity  \\
\hline
Histogram-based Outlier Score (HBOS) & Proximity  \\
\hline
 Minimum Covariance Determinant (MCD) & Linear Model \\
\hline
One-Class Support Vector Machines (OCSVM) & Linear Model\\
\hline
Connectivity-Based Outlier Factor (COF)  & Proximity\\
\hline
Principal Component Analysis (PCA) & Linear Model \\
\hline
Feature Bagging & Ensembling\\
\hline
Isolation Forest & Ensembling\\
\hline
Subspace Outlier Detection (SOD) & Proximity \\
\hline
Lightweight On-line Detector of Anomalies (LODA) & Ensembling\\
\hline
Locally Selective Combination of Parallel Outlier Ensembles (LSCP) & Ensembling \\
\hline Copula-Based Outlier Detection (CPOD) & Probabilistic \\
\hline
\end{tabular}
\label{table1}
\caption{Outlier Detection Techniques Categorization.}
\end{table}
\subsection{Proximity Based Outlier Detection Techniques}
% \cite{aggarwal2017proximity}
\subsubsection{(a) LOF : Local Outlier Factor}
Local Outlier Factor algorithm (LOF) \cite{breunig2000lof} finds out how isolated a point is with respect to the surrounding neighborhood. Each data point is given a score (LOF score of an observation is equal to the ratio of the average local density of his k-nearest neighbors, and its own local density) that tells how isolated it is based on the size of its local neighborhood. The points with the largest score are more likely to be outliers. This algorithm works well for feature spaces with low dimensionality (curse of dimensionality). It can perform well even in datasets where abnormal samples have different underlying densities. An inlier is expected to have a local density similar to that of its neighbors, while an outlier is expected to have much smaller local density.

\subsubsection{(b) KNN: K Nearest Neighbors}
K Nearest Neighbors (KNN) \cite{5070231} can be used as an anomaly detector as it has an unsupervised approach. In unsupervised learning, there is no pre-determined labeling and there is no actual learning in the process. KNN calculates k- nearest neighborhood from a data point. It can use the largest distance of the kth neighbor as the outlier score or average of all k neighbors or the median of the distance to k neighbors as the outlier score. It is entirely based upon threshold values. Data scientists decide a cutoff value beyond which all observations are called anomalies. 
12

\subsubsection{(c) CBLOF : Clustering-Based Local Outlier Factor}
The outlier detection algorithms based on the traditional approaches use all the features of the data set, thus, causing massive computational cost for the large data sets. To minimize the computational, authors first divide the large data set into meaningful clusters and then utilize concept from Local Outlier Factor (LOF) \cite{kriegel2009loop} to detect top n outliers. To identifying the physical significance of an outlier, authors proposed a measure CBLOF. Physcial significance is measured by both the size of the cluster the object belongs to and the distance between object and its closest cluster.

\subsubsection{(d) ABOD : Angle-Based Outlier detection}
Distance based algorithms fail to give quality results on high-dimensional data. Angle based approach suggested in 2008, to detect the outlier is independent of parameters. It uses the variance of the angles between the difference vectors of data objects to find outliers and also alleviates the effects of the ``curse of dimensionality" on mining high-dimensional data \cite{inproceedings}. Angle between farthest data point is less than the angle between nearer data points and hence the variance for the distant points will be lesser than the nearer points. The data point with lesser variance is considered as an outlier. Another fact to be considered is that a point A is considered as an outlier if most other points are located in similar directions.

\subsubsection{(e) HBOS : Histogram-based Outlier Score}
Histogram Based Outlier Score (HBOS) \cite{goldstein2012histogram} is a fast unsupervised, statistical and non-parametric method that can be used to detect the outliers. It assumes that all the features are independent. In case of categorical data, simple counting is used while for numerical values, static or dynamic bins are made. The height of each bin represents the density estimation. To ensure an equal weight of each feature, the histograms are normalized [0-1]. In HBOS, outlier score is calculated for each single feature of the dataset. These calculated values are inverted such that outliers have a high HBOS and inliers have a low score. HBOS include a fast computation time, scoring-based detection, and absence of a learning phase. It assumes feature independence and hence is ineffective in case of high dimensionality.

\subsubsection{(f) SOD: Subspace Outlier Detection}
The inliers and outliers can be distinguished by the fact that the inliers fit in to the subspace spanned by a set of reference points. If the data is d-dimensional then an axis-parallel hyperplane with dimensions less than d, will be the reference point. In the subspace, variance of the reference points is high. If a point is in the subspace of this hyperplane, it will be an inlier but if it deviates significantly then it is an outlier in subspace perpendicular to that hyperplane and its variance in perpendicular subspace is low \cite{kriegel2009outlier}. 

\subsubsection{(g) COF : Connectivity-Based Outlier Factor}
Connectivity based outlier (COF) \cite{tang2002enhancing} algorithm assigns degree of outlier to each data point which is called connectivity based outlier factor (COF) of the data point. COF is the ratio of average chaining distance of data point and the average of average chaining distance of k nearest neighbor of the data point. High COF value shows high probability of being an outlier.

\subsection{Linear Model Based Outlier Detection Techniques}
\subsubsection{(a) MCD: Minimum Covariance Determinant}
The minimum covariance determinant (MCD) \cite{pesch1999computation,hubert2010minimum} finds its application in financial fields, medical domains, image analysis and many more. MCD is a very basic and robust method developed in 1999 and is used to develop many other robust multivariate techniques like PCA, multiple regression, etc. MCD is generally used to detect outliers in multivariate data. Assuming the underlying distribution as elliptical symmetric unimodal distribution with unknown parameters mean and covariance, a robust tolerance ellipse (on dataset) based on the robust distances is much smaller than other possible ellipses and encloses only regular data points. As the robust distances are not sensitive to masking effect, they are used to separate out the outliers. This robust ellipse gives the MCD estimators mean and covariance (also called as minimum covariance).
Main properties of MCD are affine equivariance, breakdown value, and influence function. Recent extensions of MCD are fast deterministic algorithm and second is used for high‐dimensional data. Fast deterministic algorithm is highly robust and almost affine equivariant to MCD.

\subsubsection {(b) OCSVM : One-Class Support Vector Machines}
One-Class Support Vector Machines (OCSVM) \cite{bounsiar2014one} predicts outliers for both classification and regression dataset. OCSVM algorithm captures the density of the majority class and for the points on the extremes of the density function as outliers. OCSVM computes a binary function that captures regions in input space, where the probability density is high. When OneClassSVM overfits, the scores of abnormality of the samples as a good estimator should assign similar scores to all the samples.

\subsubsection{(c) PCA : Principal Component Analysis}
Principal component analysis (PCA) \cite{saha2016application} is an unsupervised algorithm that does not require the data to be labelled as ``outlier" or ``inlier". It is generally used for dimensinality reduction, but the main essence of its working lies in the fact that it is used to analyze the inherent structure of the data. When PCA receives N samples and D dimensions(variables), each variable represents one co-ordinate axis. PCA reduces these dimensions by finding different or alternative variable (coordinate axes). Principle Components are the axes to which the dimensions D are reduced to. First Principal Component (PC1) is the line in the d-dimensional space that passes through the average of the data points and best approximates the data in the least square sense. It represents the maximum of the total variance in the observed variables. Second Principal Component (PC2) is orthogonal to PC1 in k-dimensional space. This line also passes through the average point and improves the approximation of the X-data to the possible extent . This helps to find out the points that are away from other data points and are called as outliers. In PCA, the original variables are lost and thus new Principle components are not as readable and interpretable as original features. Before applying PCA, data needs to be standardized and scaled. PCA in itself is computationally expensive, but help in removing outliers and improve visualization.

\subsection{Ensembling Based Outlier Detection Techniques}
% \cite{zimek2014ensembles}
\subsubsection{(a) Feature Bagging}
Every outlier detection algorithm uses a small subset of features to detect the outliers, like distance or angle between data points or the cluster density, etc. After that the algorithms assign an outlier score to all the data points that correspond to their probability of being outliers. As a result, each algorithm identifies different set of outliers \cite{10.1145/1081870.1081891}.
Feature bagging techniques combines results from several algorithms that are applied and combine the outlier scores computed by the individual algorithms are then combined in order to find the better quality outliers. This approach is generally used for high dimensional and noisy data.

\subsubsection{(b) Isolation Forest}
The Isolation Forest algorithm \cite{liu2008isolation} is based on binary decision trees. It picks a feature from the feature space and a random split value. The trees are made with these chosen features and splits. To build the forest, a tree ensemble is made by  averaging all the trees in the forest. Then for prediction, it compares an observation against that splitting value in a ``node". The number of ``splittings" is called as ``path length". Outliers will have shorter path lengths than the rest of the observations. Isolation Forest has the advantage that there is no scaling of the values and is also effective when value distributions can not be assumed. This makes it robust and optimizable. If not correctly optimized, training time can be very long and computationally expensive.

\subsubsection{(c) LODA : Lightweight On-line Detector of Anomalies}
Lightweight On-line Detector of Anomalies (LODA) \cite{10.1007/s10994-015-5521-0} is particularly useful when huge data is processed in real time. LODA is not only fast and accurate but also able to operate and update itself on data with missing variables. LODA can identify features in which the given sample deviates from the majority, which basically finds out the cause of anomaly. The LODA algorithm constructs an ensemble of T one-dimensional histogram density estimators. LODA is a collection of weak classifiers can result in a strong classifier.

\subsubsection{(d) LSCP : Locally Selective Combination of Parallel Outlier Ensembles}
In unsupervised outlier ensembles, there are no labels for "outliers" and "inliners". Therefore, it is challenging to find a reliable way of selecting competent base detectors and stability during model combination. Traditional unsupervised combination algorithms in parallel ensembles are often generic and global like averaging, maximization, weighted averaging, etc but they do not consider locality. LSCP \cite{zhao2019lscp} define a local region around a datapoint using the concept of its nearest neighbors in randomly selected feature subspaces. The top-performing base detectors in this local region are selected and ensembled to make the final model.

\subsection{Probabilistic Based Outlier Detection Techniques}
\subsubsection{(a) COPOD: Copula-Based Outlier Detection}
COPOD is parameter-free, and highly interpretable outlier detection technique proposed in \cite{li2020copod}. This work is inspired by copulas for modeling multivariate data distribution. COPOD first constructs an empirical copula, and then uses it to predict tail probabilities of each given data point to determine its level of extremeness. This is an efficient algorithm that scales well in high dimensional settings. Unlike proximity based models that require pairwise distance calculation \cite{zhao2018xgbod,li2020sync} or learning based models that require training, COPOD incurs low computational overhead.

\subsection{Other techniques}
There are several other outlier techniques present in the dataset such as (a) Linear model for deviation Detection (LMDD) from linear model categorization \cite{hand2014data}, (b) Fast outlier detection using the local correlation integral (LOCI) \cite{papadimitriou2003loci}, Average kNN, Median kNN, Rotation-based Outlier Detection \cite{zhu2020knn} from proximity based categorization, (c) Fast Angle-Based Outlier Detection using approximation \cite{pham2018l1}, Median Absolute Deviation \cite{leys2013detecting}, Stochastic Outlier Selection from probabilistic based categorization \cite{jero}, and (d) Extreme Boosting Based Outlier Detection \cite{zhao2018xgbod}, Lightweight On-line Detector of Anomalies from Ensemble based categorization. There is another category called as Neural Networks which exist in literature, however we are not comparing those because of high computational cost. Some of the existing techniques which is based on Neural Network \cite{chen2017outlier,liu2019generative} concept are Fully connected AutoEncoder (use reconstruction error as the outlier score), Variational AutoEncoder (use reconstruction error as the outlier score), Single-Objective Generative Adversarial Active Learning, and Multiple-Objective Generative Adversarial Active Learning.

\begin{table} [t]
\tiny
 \begin{minipage}{.45\linewidth}
\begin{tabular}{ |c|c|c|c|c| } 
 \hline
 Dataset & \# samples & \# dim&  \# outliers &\% outliers\\ 
 \hline
 \hline
Annthyroid & 7200 & 6 & 534 & 7.42\\ 
\hline
Arrhythmia & 452 & 274 & 66 & 15 \\ 
\hline
BreastW & 683	& 9	& 239 & 35 \\
\hline
Cardio & 1831 &	21 & 	176 & 9.6\\
\hline
Glass & 214&	9&	9 & 4.2\\
\hline
Ionosphere & 351 &	33&	126 & 36 \\
\hline
Letter & 1600 &	32 &	100 & 6.25 \\
\hline
Lympho  & 148 &	18 &	6 & 4.1 \\
\hline
Mnist & 7603 & 100 &	700  & 9.2 \\
\hline
Musk& 3062	& 166	& 97 & 3.2 \\
\hline
Optdigits & 5216 &	64 &	150 & 3 \\
\hline
Pendigits & 6870 &	16	 &156 & 2.27 \\
\hline
Pima & 768 & 8& 268 & 35 \\
\hline
Satellite & 6435 & 36 & 2036 & 32 \\
\hline
Satimage-2 & 5803	& 36 &	71 & 1.2\\
\hline
Thyroid & 3772 &	6 &	93 & 2.5\\
\hline
Vertebral & 240	& 6 & 	30 & 12.5\\
\hline
Vowels & 1456 & 12	& 50 & 3.4\\
\hline
WBC & 278 & 30	& 21 & 5.6\\
\hline
Wine & 129 & 13	& 10 &  7.7\\
\hline
\end{tabular}
\begin{center}
\caption{Dataset details to measure Precision \& Recall.}
 \end{center}
\end{minipage}
\hspace{0.4 cm}
 \begin{minipage}{.55\linewidth}
\begin{tabular}{ |c|c|c|c|c| } 
 \hline
Dataset &  \# samples &  \# dim &  \# categorical & \# numerical\\
 &   &   &   cols & cols\\
 \hline
 \hline
 Annthyroid & 7200 & 21 & 14 & 7\\ 
\hline
 Arrhythmia & 452 & 278 & 72  & 206\\ 
\hline
BreastW & 699	& 10 & 1 & 10 \\
\hline
Glass  & 214&	10& 1 & 10\\
\hline
Heart & 267 &	44 & 1 & 44\\
\hline
Ionosphere & 351 &	34 & 1 & 34\\
\hline
Letter & 20000 & 16 & 1 & 16\\
\hline
Lympho  & 148 &	18 & 19 & 0\\
\hline
Optdigits & 5620 &	64 & 0 & 65\\
\hline
Pendigits & 10992 &	16 & 0 & 17\\
\hline
Satellite & 6435 & 36 & 1 & 36\\
\hline
Vertebral & 310	& 6 & 1 & 6\\
\hline
WBC & 569 & 32& 1 & 32\\
\hline
Wine & 178 & 13& 1 & 13\\
\hline
\end{tabular}
\begin{center}
\caption{Dataset details to measure Model Classification.}
 \end{center}
\end{minipage}
\end{table}

\section{Evaluation}
\label{res}
\subsection{Evaluation Metrics}
We have conducted two types of experiments. In the first experiment, we compare the performance of different outlier detection algorithms (discussed in Section 2) using precision and recall metric. This helps to decide which algorithm is detecting the outliers more correctly. In the second experiment, we measure the effect of outlier removal on model classification accuracy. This kind of analysis helps in understanding that which algorithm is detecting the points as outliers which can affect the ML model performance most.
We used three ML classifiers Logistic Regression, Decision Trees and Random Forest to measure this effect.

% \begin{table}[t]
% \scriptsize
% \begin{center}
% \begin{tabular}{ |c|c|c|c|c| } 
%  \hline
% Dataset & \# samples & \# dim & \# categorical columns & \# numerical columns \\ 
%  \hline
%  \hline
%  Annthyroid & 7200 & 21 & 0 & 22\\ 
% \hline
%  Arrhythmia & 452 & 278 & 0 & 279\\ 
% \hline
% BreastW & 699	& 10 & 1 & 10 \\
% \hline
% Glass  & 214&	10& 1 & 10\\
% \hline
% Heart & 267 &	44 & 0 & 45\\
% \hline
% Ionosphere & 351 &	34 & 1 & 34\\
% \hline
% Letter Recognition & 20000 & 16 & 0 & 17\\
% \hline
% Lympho  & 148 &	18 & 0 & 19\\
% \hline
% Optdigits & 5620 &	64 & 0 & 65\\
% \hline
% Pendigits & 10992 &	16 & 0 & 17\\
% \hline
% Satellite & 6435 & 36 & 0 & 37\\
% \hline
% Vertebral & 310	& 6 & 0 & 7\\
% \hline
% WBC & 569 & 32& 0 & 33\\
% \hline
% Wine & 178 & 13& 0 & 14\\
% \hline
% \end{tabular}
% \end{center}
% \caption{Dataset Details for Model Classification.}
% \end{table}
\subsection{Dataset Details}
For comparison of different outlier detector techniques, we used multi-dimensional point-outlier detection datasets \cite{outD}. It openly provide access to a large collection of outlier detection datasets with ground truth. Details of 20 dataset under observation are provided in Table 2.

For the task of Model Performance comparison before and after outlier removal, the datastes corresponding to datasets from  multi dimensional point-outlier detection datasets are taken from UCI repository \cite{UCI}. We can't use multi dimensional point-outlier detection datasets directly because these datasets does not have class label information. We are able to get 14 datasets corresponding to multi dimensional point-outlier detection from UCI repository. Details of 14 dataset under observation are provided in Table 3.

\subsection{Configuration Details}
% In this paper, two tasks are performed and the results are recorded accordingly. The first task is to find out the performance of different outlier detection models on the ODDs(Outlier Detection Dataset). In ODDS, there is an open access to a large collection of Multi-dimensional point datasets with ground truth . The data is from multiple domains, from satellite data to Cardio data and from optical digit recognition to wine and glass classification data.
% Fifteen Different outlier detection models developed from 1999 to 2020 are used to find out how they perform on different kinds of datasets. 
'Contamination' is the proportion of the most isolated points that will be considered as outliers. In this experiment, contamination value is set to 0.1 for all the models so that better comparison can be done. Increasing the contamination will increase the number of predicted outliers but may not increase the precision and the recall score as the true positives may remain same or may even decrease. For model evaluation, we first converted the categorical values are converted to the numeric values using lambda categorical coding techniques. We used Logistic Regression, Decision Trees and Random Forests for training. Results are reported averaged over 3 folds. In each fold, grid search is performed to find the best parameters like criteria-gini and entropy, different depths in decision trees, C values in Logistic Regression. This is the baseline modelling done with the full data set. After this, 15 outlier Detection Models (discussed in Section 2) are applied one by one to remove the outliers from training, and classifier performance is reported.

% \begin{table}[t]
% \begin{center}
% \begin{tabular}{ |c|c|c|c| } 
%  \hline
%  Dataset & Logistic Regression & Decision Tree &  Random Forest\\ 
%  \hline
%  \hline
% Annthyroid & 95.1 & 99.5 & 99.4\\ 
% \hline
%  Arrhythmia & 71.0 & 62.8 & 71.3\\ 
% \hline
% BreastW & 95.7 &94.1 & 96.3 \\
% \hline
% Glass  & 59.2 & 68.3 & 76.5 \\
% \hline
% Heart & 78.3 & 76.8 & 82.4 \\
% \hline
% Ionosphere & 83.2 & 85.4 & 91.1 \\
% \hline
% Letter Recognition & 76.0 & 87.1 & 95.3\\
% \hline
% Lympho  & 83.0 & 72.9 & 85.0\\
% \hline
% Optdigits & 96.5 &	89.5 & 98.0 \\
% \hline
% Pendigits & 93.5 & 96.0 & 99.4 \\
% \hline
% Satellite & 79.2 & 85.8 & 91.1\\
% \hline
% Vertebral & 83.6 & 77.8 & 80.3\\
% \hline
% WBC & 94.3 & 93.0 & 96.7 \\
% \hline
% Wine & 93.8 & 95.5 & 98.3 \\
% \hline
% \end{tabular}
% \end{center}
% \caption{Classifier performance on Raw Data (without outlier Removal)}
% \end{table}

\begin{sidewaystable}
% \begin{table}
\scriptsize
\caption{Precision Analysis of Outlier Algorithms.}
\begin{tabular}{ |c|c|c|c|c|c|c|c|c|c|c|c|c|c|c|c|c|c|c| } 
 \hline
 \textbf{Dataset}  & \textbf{ABOD} &	\textbf{COPOD} &	\textbf{CBLOF} &	\textbf{F.Bagging} & \textbf{HBOS} & \textbf{IForest} &	\textbf{KNN} & \textbf{Avg KNN} &	\textbf{LOF} &	\textbf{MCD} &	\textbf{OCSVM} &	\textbf{PCA} &	\textbf{LSCP} &	\textbf{COF} &	\textbf{SOD} &	\textbf{LODA}\\
 \hline
\textbf{annthyroid}&	0.000&	0.222&	0.300&	0.206&	0.298&	0.353&	0.317&	0.322&	0.333&	\textbf{0.564}&	0.128&	0.281&	0.344&	0.328&	0.328&	0.069\\
 \hline
\textbf{arrhythmia}&	0.435&	0.652&	\textbf{0.783}&	0.652&	0.478&	0.565&	0.565&	0.522&	0.609&	0.696&	0.000&	0.522&	0.609&	\textbf{0.783}&	0.435&	0.457\\
 \hline
\textbf{breastw}&	0.000&	\textbf{1.000}&	\textbf{1.000}&	0.000&	0.971&	\textbf{1.000}&\textbf{	1.000}&	0.943&	0.200&	0.971&	0.943&	\textbf{1.000}&	0.086&	0.000&	0.800&	\textbf{1.000}\\
 \hline
\textbf{cardio}&	0.272&	0.609&	\textbf{0.641}&	0.228&	0.554&	0.620&	0.457&	0.413&	0.239&	0.391&	0.533&	0.587&	0.272&	0.207&	0.348&	0.574\\
 \hline
\textbf{glass}&	0.091&	0.091&	0.091&	\textbf{0.182}&	0.091&	0.091&	0.091&	0.091&	0.091&	0.000&	0.091&	0.091&	\textbf{0.182}&	0.091&	\textbf{0.182}&	0.045\\
 \hline
\textbf{ionosphere}&	\textbf{1.000}&	0.944&	\textbf{1.000}&	\textbf{1.000}&	0.222	&\textbf{1.000}&\textbf{1.000}&	\textbf{1.000}&	\textbf{1.000}&	\textbf{1.000}&	\textbf{1.000}&	\textbf{1.000}&	\textbf{1.000}&	\textbf{1.000}&	\textbf{1.000}&	\textbf{1.000}\\
 \hline
\textbf{letter}&	0.488&	0.038&	0.188	&0.450&	0.075&	0.063&	0.423&	\textbf{0.525}&	0.425&	0.188&	0.488&	0.088&	0.463&	0.450&	\textbf{0.525}&	0.119\\
 \hline
\textbf{lympho}&	0.250&	0.625&	0.250&	0.625&	\textbf{0.750}&	0.625&	0.667&	0.500&	0.625&	0.250&	0.500&	0.625&	\textbf{0.750}&	0.375&	0.375&	0.067\\
 \hline
\textbf{mnist}&	0.310&	0.236&	0.404&	0.373&	0.181&	0.370&	\textbf{0.491}&	0.451&	0.360&	0.310&	0.000&	0.444&	0.365&	0.333&	0.249&	0.310\\
 \hline
\textbf{musk}&	0.013&	0.325&	\textbf{0.630}&	0.019&\textbf{	0.630}&	\textbf{0.630}&	0.032&	0.006&	0.019&	\textbf{0.630}&	0.000&	\textbf{0.630}&	0.019&	0.084&	0.045&	0.293\\
 \hline
\textbf{optdigits}&	0.034	&0.015&	0.057&	0.077&\textbf{0.184}&	0.034&	0.023&	0.008&	0.077&	0.000&	0.023&	0.000&	0.069&	0.065&	0.034&	0.006\\
 \hline
\textbf{pendigits}&	0.052&	0.166&	\textbf{0.334}&	0.029&	0.206&	0.198&	0.070&	0.061&	0.029&	0.012&	0.215&	0.221&	0.038&	0.052&	0.049&	0.189\\
 \hline
\textbf{pima}&	0.436&\textbf{0.692}&	0.590	&0.487&	0.615&	0.590&	0.513&	0.487&	0.513&	0.590&	0.256&	0.538&	0.513&	0.436&	0.462&	0.403\\
 \hline
\textbf{satellite}&	0.503&	0.882&\textbf{1.000}&	0.494&	0.953&	0.953&	0.758&	0.720&	0.503&	0.975&	0.280&	\textbf{1.000}&	0.516&	0.565&	0.571&	0.963\\
 \hline
\textbf{satimage-2}&	0.093&	0.213&	0.241&	0.041&	0.210&	0.234&	0.131&	0.127&	0.041&	\textbf{0.244}&	0.007&	0.220&	0.052&	0.062&	0.089&	0.117\\
 \hline
\textbf{thyroid}&	0.000&	0.180&	0.228&	0.143&	0.354&	0.344&	0.270&	0.259&	0.180&	\textbf{0.444}&	0.122&	0.286&	0.175&	0.132&	0.185&	0.111\\
 \hline
\textbf{vertebral}&	0.000&	0.000&	0.000&	0.000&	0.000&	0.000&	0.000&	0.000&	0.000&	0.000&	\textbf{0.167}&	0.000&	0.000&	0.083&	0.000&	0.000\\
 \hline
\textbf{vowels}&	0.521&	0.027&	0.164&	0.315&	0.123&	0.178&	0.493&	0.493&	0.315&	0.041&	0.192&	0.123&	0.288&	\textbf{0.534}&	0.438&	0.082\\
 \hline
\textbf{wbc}&	0.316&	\textbf{0.737}&	0.421&	0.474&	0.632&	0.474&	0.579&	0.474&	0.474&	0.474&	0.579&	0.526&	0.474&	0.579&	0.474&	0.395\\
 \hline
\textbf{wine}&	0.571&	0.429&\textbf{	1.000}&	\textbf{1.000}&	0.000&	0.000&\textbf{	1.000}&	0.857&\textbf{	1.000}&	0.714&	0.143&	0.143&\textbf{	1.000}&	\textbf{1.000}&	0.571&	0.538\\
 \hline
\textbf{Avg Precision} & 0.269&	0.404&\textbf{0.466}&	0.340&	0.376&	0.416&	0.444&	0.413&	0.352&	0.425&	0.283&	0.416&	0.361&	0.358&	0.358&	0.337\\
\hline
 
\end{tabular}

\caption{Recall Analysis of Outlier Algorithms.}
\begin{tabular}{ |c|c|c|c|c|c|c|c|c|c|c|c|c|c|c|c|c|c|c| } 
 \hline
  \textbf{Dataset}  & \textbf{ABOD} &	\textbf{COPOD} &	\textbf{CBLOF} &	\textbf{F.Bagging} & \textbf{HBOS} & \textbf{IForest} &	\textbf{KNN} & \textbf{Avg KNN} &	\textbf{LOF} &	\textbf{MCD} &	\textbf{OCSVM} &	\textbf{PCA} &	\textbf{LSCP} &	\textbf{COF} &	\textbf{SOD} &	\textbf{LODA}\\
 \hline
 \hline
\textbf{annthyroid} & 0.000&	0.150&	0.202&	0.139&	0.202&	0.238&	0.213&	0.217&	0.225&	\textbf{0.380}&	0.086&	0.189&	0.232&	0.221&	0.221&	0.094\\
\hline
\textbf{Arrhythmia} & 0.152&	0.227&	0.273&	0.227&	0.167&	0.197&	0.197&	0.182&	0.212&	0.242&	0.000&	0.182&	0.212&	0.273&	0.152&	\textbf{0.318}\\
\hline
\textbf{BreastW} & 0.000	&0.146&	0.146&	0.000&	0.142&	0.146&	0.134&	0.138&	0.029&	0.142&	0.138&	0.146&	0.013&	0.000&	0.117&	\textbf{0.289}\\
\hline
\textbf{cardio}&	0.142	&0.318&	0.335&	0.119&	0.290&	0.324&	0.239&	0.216&	0.125&	0.205&	0.278&	0.307&	0.142&	0.108&	0.182&	\textbf{0.597}\\
\hline
\textbf{glass}&	0.111&	0.111&	0.111&	\textbf{0.222}&	0.111&	0.111&	0.111&	0.111&	0.111&	0.000&	0.111&	0.111&	\textbf{0.222}&	0.111&	\textbf{0.222}&	0.111\\
\hline
\textbf{ionosphere}&	0.143&	0.135&	0.143&	0.143&	0.032&	0.143&	0.143&	0.143&	0.143&	0.143& 0.143&	0.143&	0.143&	0.143&	0.143&	\textbf{0.278}\\
\hline
\textbf{letter}&	0.390&	0.030&	0.150&	0.360&	0.060&	0.050&	0.330&	\textbf{0.420}&	0.340&	0.150&	0.390&	0.070&	0.370&	0.360&	\textbf{0.420}&	0.190\\
\hline
\textbf{lympho}&	0.333&	0.833&	0.333&	0.833&	\textbf{1.000}&	0.833&	0.667&	0.667&	0.833&	0.333&	0.667&	0.833&	\textbf{1.000}&	0.500&	0.500&	0.167\\
\hline
\textbf{mnist} &	0.169&	0.129&	0.220&	0.203&	0.099&	0.201&	\textbf{0.267}&	0.246&	0.196&	0.169&	0.000&	0.241&	0.199&	0.181&	0.136&	0.337\\
\hline
\textbf{musk}&	0.021&	0.515&	\textbf{1.000}&	0.031&	\textbf{1.000}&	\textbf{1.000}&	0.052&	0.010&	0.031&	\textbf{1.000}&	0.000&\textbf{1.000}&	0.031&	0.134&	0.072&	0.928\\
\hline
\textbf{optdigits}&	0.060&	0.027&	0.100&	0.133&	\textbf{0.320}&	0.060&	0.040&	0.013&	0.133&	0.000&	0.040&	0.000&	0.120&	0.113&	0.060&	0.020\\
\hline
\textbf{pendigits}&	0.115&	0.365&	0.737&	0.064&	0.455&	0.436&	0.154&	0.135&	0.064&	0.026&	0.474&	0.487&	0.083&	0.115&	0.109&	\textbf{0.833}\\
\hline
\textbf{pima}&	0.063&	0.101&	0.086&	0.071&	0.090&	0.086&	0.075&	0.071&	0.075&	0.086&	0.037&	0.078&	0.075&	0.063&	0.067&	\textbf{0.116}\\
\hline
\textbf{satellite}&	0.080&	0.139 & 0.158	& 0.078 &	0.151&	0.151	&0.120&	0.114	&0.080&	0.154&	0.044&	0.158&	0.082&	0.089&	0.090&	\textbf{0.305}\\
\hline
\textbf{satimage-2} &	0.380&	0.873&	0.986&	0.169&	0.859&	0.958&	0.535&	0.521&	0.169&	\textbf{1.000}&	0.028&	0.901&	0.211&	0.254&	0.366&	0.958\\
\hline
\textbf{thyroid}&	0.000&	0.366&	0.462&	0.290&	0.720&	0.699&	0.548&	0.527&	0.366&	\textbf{0.903}&	0.247&	0.581&	0.355&	0.269&	0.376&	0.452\\
\hline
\textbf{vertebral}&	0.000&	0.000&	0.000&	0.000&	0.000&	0.000	&0.000&	0.000&	0.000&	0.000&	\textbf{0.067}&	0.000&	0.000&	0.033&	0.000&	0.000\\
\hline
\textbf{vowels}&	0.760	&0.040	&0.240&	0.460&	0.180&	0.260&	0.720&	0.720&	0.460&	0.060&	0.280&	0.180&	0.420&	\textbf{0.780}&	0.640&	0.240\\
\hline
\textbf{wbc}	&0.286&	0.667&	0.381&	0.429&	0.571&	0.429&	0.524&	0.429&	0.429&	0.429	&0.524&	0.476&	0.429&	0.524&	0.429&	\textbf{0.714}\\
\hline
\textbf{wine} &	0.400 &	0.300&	\textbf{0.700}&	\textbf{0.700}&	0.000&	0.000&	\textbf{0.700}&	0.600&	\textbf{0.700}&	0.500&	0.100&	0.100&	\textbf{0.700}&	\textbf{0.700}&	0.400&	\textbf{0.700}\\
\hline
\textbf{Avg Recall} & 0.180&	0.274&	0.338&	0.234&	0.322&	0.316&	0.288&	0.274&	0.236&	0.296&	0.183&	0.309&	0.252&	0.249&	0.235&	\textbf{0.382}\\
\hline
\end{tabular}
\end{sidewaystable}

\begin{sidewaystable}
% \begin{table}
\scriptsize
\caption{Performance of Random Forest on different outlier algorithms.}
\begin{tabular}{ |c|c|c|c|c|c|c|c|c|c|c|c|c|c|c|c|c|c|c|c| } 
 \hline
 \textbf{Dataset} & \textbf{Original} & \textbf{ABOD} &	\textbf{COPOD} &	\textbf{CBLOF} &	\textbf{F.Bagging} & \textbf{HBOS} & \textbf{IForest} &	\textbf{KNN} & \textbf{Avg KNN} &	\textbf{LOF} &	\textbf{MCD} &	\textbf{OCSVM} &	\textbf{PCA} &	\textbf{LSCP} &	\textbf{COF} &	\textbf{SOD} &	\textbf{LODA}\\
 \hline
 \hline
\textbf{annthyroid} & 99.4 & \textbf{99.44}&	99.36&	\textbf{99.46}&	\textbf{99.53}&	\textbf{99.47}&	\textbf{99.44}&	99.40&	99.40&	99.39&	98.75&	\textbf{99.43}&	99.31&	99.40&	\textbf{99.43}&	\textbf{99.46}&	99.40\\
\hline
\textbf{Arrhythmia}& 71.3 & \textbf{71.39}&	68.07&	70.06&	66.74&	67.84&	67.4&	66.29&	66.96&	67.85&	67.85&	\textbf{71.4}&	68.95&	68.07&	68.51&	70.51&	68.07 \\ 
\hline
\textbf{BreastW} & 96.3 & \textbf{97.13}	&\textbf{96.42}&	96.13&	\textbf{96.85}&	95.99&	96.13&	96.28& \textbf{96.42}&	96.28 & \textbf{96.56}	& 96.28 &	96.27&	\textbf{96.42}&	\textbf{96.99}&	95.99&	\textbf{96.71}\\
\hline
\textbf{Glass} & 76.5 & \textbf{77.0}&	73.71&	74.18&	75.12&	75.59&	73.71&	75.59&	72.3&	69.48&	75.12&	\textbf{77.0}&	75.12&	74.18&	73.71&	73.24&	\textbf{78.4}\\
\hline
\textbf{Heart}& 82.4 & 82.02&	82.02&	80.15&	\textbf{82.42}&	80.15&	78.65&	76.03&	79.40&	79.78&	80.52&	80.52&	\textbf{82.42}&	81.27&	80.90&	80.90&	79.78\\
\hline
\textbf{Ionosphere} & 91.1 & \textbf{91.16}& \textbf{92.58}&	\textbf{91.72}&	90.01&	\textbf{91.44} &	90.87&	\textbf{91.73}&	  90.87&	89.45&	\textbf{91.15}&	89.44&	\textbf{92.01}&	  90.30&	90.87&	\textbf{92.01}&	    90.59\\
\hline
\textbf{Letter}  & 95.3 & \textbf{96.01}&	95.28&	94.93&	94.59&	\textbf{95.50}&	\textbf{95.49}&	93.91&	93.71&	94.51&	94.59&	94.01&	\textbf{95.30}&	94.57&	94.95&	94.75&	\textbf{95.47}\\
\hline
\textbf{Lympho}  & 85.0 & 	82.31&	81.63&	84.35&	83.67&	82.99&	83.67&	84.35&	80.95&	\textbf{85.03}&	\textbf{85.03}&	83.67&	81.63&	82.99&	84.35&	83.67&	\textbf{86.39}\\
\hline
\textbf{Optdigits} & 98.0 & 97.6&	97.76&	96.81&	97.46&	\textbf{98.01}&	97.42&	96.76&	96.98&	97.51&	97.56&	\textbf{98.04}&	97.62&	97.46&	97.26&	97.46&	97.51\\
\hline
\textbf{Pendigits} & 99.4 & \textbf{99.63}&	\textbf{99.51}&	99.20&	99.27&	\textbf{99.65}&	99.26&	99.29&	\textbf{99.60}&	99.38&	\textbf{99.65}&	\textbf{99.78}&	\textbf{99.65}&	99.39&	\textbf{99.47}&	\textbf{99.67}&	\textbf{99.53} \\
\hline
\textbf{Satellite}  & 91.1 & 90.63&	\textbf{91.16}&	89.23&	90.3&	\textbf{91.17}&	90.89&	89.43&	89.76&	90.16&	89.34&	90.96&	\textbf{91.11}&	90.15&	90.37&	90.43&	\textbf{91.39}\\
\hline
\textbf{Vertebral} & 80.3 &\textbf{85.82}&	\textbf{85.48}&	\textbf{84.52}&	\textbf{84.52}&	\textbf{86.13}&	\textbf{87.75}&	\textbf{85.17}&	\textbf{85.49}&	\textbf{86.13}&	\textbf{84.85}&	\textbf{85.48}&	\textbf{85.81}&	\textbf{85.50}&	\textbf{84.84}&	\textbf{86.13}&	\textbf{84.52} \\
\hline
\textbf{WBC} & 96.7 & 95.07	&95.42&	95.95&	94.71&	95.42&	94.54&	95.42&	95.24&	94.89&	95.42	&95.77&	95.42&	93.13&	94.89&	95.25&	95.77\\
\hline
\textbf{Wine} & 98.3 & \textbf{98.31}&	97.75&	96.07&	97.75&	97.20&	97.75&	\textbf{98.31}&	97.75&	97.75	&97.19&\textbf{	98.88}&	\textbf{98.31}&	\textbf{98.87}&	97.75&	97.75&	\textbf{98.87}\\
\hline
\end{tabular}

\caption{Performance of Decision Tree on different outlier algorithms.}
\begin{tabular}{|c|c|c|c|c|c|c|c|c|c|c|c|c|c|c|c|c|c|c| } 
 \hline
 \textbf{Dataset} & \textbf{Original} & \textbf{ABOD} &	\textbf{COPOD} &	\textbf{CBLOF} &	\textbf{F.Bagging} & \textbf{HBOS} & \textbf{IForest} &	\textbf{KNN} & \textbf{Avg KNN} &	\textbf{LOF} &	\textbf{MCD} &	\textbf{OCSVM} &	\textbf{PCA} &	\textbf{LSCP} &	\textbf{COF} &	\textbf{SOD} &	\textbf{LODA}\\
 \hline
 \hline
 \textbf{Annthyroid}& 99.5 & \textbf{99.57}	&99.43&	\textbf{99.53}&	\textbf{99.53}&	99.47&	\textbf{99.51}&	99.40&	99.43&	99.47&	99.08&	\textbf{99.53}&	99.40&	99.42&	99.21&	99.39&	\textbf{99.51}\\ 
\hline
 \textbf{Arrhythmia}& 62.8 & 61.86&	62.53&	\textbf{64.74}&	\textbf{65.63}&	\textbf{64.97}&	62.08&	\textbf{65.63}&	\textbf{65.19}&	\textbf{66.08}&	\textbf{63.64}&	\textbf{63.86}&	\textbf{66.52}&	\textbf{65.18}&	\textbf{65.41}&	\textbf{64.53}&	\textbf{64.53} \\ 
\hline
\textbf{BreastW} & 94.1 &\textbf{94.56}&\textbf{94.27}&	\textbf{94.12}&	93.98&	92.55&	92.98&	93.84&	93.13&	93.7&	93.98&	93.41& \textbf{94.13}&	93.84&	93.98&	91.69&	93.7\\
\hline
\textbf{Glass} & 68.3 & \textbf{69.01}&	64.32&	\textbf{68.54}&	66.2&	63.38&	65.26&	66.67&	63.38&	67.14&	\textbf{68.54}&	61.5&	68.08&	\textbf{71.83}&	66.2&	67.61&	64.32\\
\hline
\textbf{Heart} & 76.8 & \textbf{79.40}&	74.16&	\textbf{77.15}&	76.8&	\textbf{78.28}&	\textbf{78.65}&	74.53&	75.28&	\textbf{79.78}&	73.03&	\textbf{79.78}&	74.53&	\textbf{77.90}&	\textbf{79.03}&	\textbf{77.90}&	\textbf{78.28}\\
\hline
\textbf{Ionosphere} & 85.4 & 83.73&	\textbf{88.58}&	82.59&\textbf{	86.29}&	\textbf{87.73}&	85.16&	\textbf{89.44}&	\textbf{89.15}&	\textbf{89.16}&	\textbf{86.02}&\textbf{	86.30}&	\textbf{88.01}&	\textbf{86.88}&	\textbf{88.59}&	83.45&\textbf{	86.59}\\
\hline
\textbf{Letter}& 87.1 & 87.0&0	85.61&	84.05&	85.38&	85.63&	85.13&	85.04&	84.80&	85.20&	85.13&	84.36&	85.43&	85.33&	85.25&	85.50&	85.22\\
\hline
\textbf{Lympho}& 72.9  & \textbf{74.15}	&\textbf{79.59}&	\textbf{80.27}&	\textbf{81.63}&	\textbf{78.91}&	\textbf{80.27}&	\textbf{79.59}&	\textbf{80.27}&	\textbf{80.95}&	\textbf{78.91}&	\textbf{79.59}&	\textbf{80.27}&	\textbf{83.6}7&	\textbf{79.59}&	\textbf{80.95}&	\textbf{82.31}\\
\hline
\textbf{Optdigits}& 89.5 & 89.0&	88.67&	87.83&	89.36&	\textbf{90.11}&	88.24&	88.56&	89.02&	89.02&	88.65&	\textbf{90.12}&	88.7&	88.9&	88.84&	89.2&	88.63 \\
\hline
\textbf{Pendigits} & 96.0 &\textbf{ 100}&	\textbf{100}&	\textbf{100}&	\textbf{99.86}&	\textbf{100}&	\textbf{100}&	\textbf{100}&	\textbf{100}&	\textbf{99.95}&	\textbf{100}&	\textbf{100}&	\textbf{100}&	\textbf{99.78}&	\textbf{100}&	\textbf{100}&	\textbf{100}\\
\hline
\textbf{Satellite} & 85.8 & 84.54&	84.82&	84.1&	84.93&	85.21&	84.86&	84.27&	84.12&	84.37&	84.06&	85.14&	85.24&	85.3&	84.71&	84.88&	84.88 \\
\hline
\textbf{Vertebral} & 77.8 & \textbf{82.89}&\textbf{	79.02}&	\textbf{81.93}&	\textbf{79.02}&	\textbf{78.39}&	\textbf{79.67}&	77.75&	\textbf{80.02}&	77.08&	\textbf{79.05}&	\textbf{80.32}&	\textbf{79.02}&	\textbf{79.67}&	\textbf{80.97}&	\textbf{78.71}&	\textbf{79.02 }\\
\hline
\textbf{WBC} & 93.0 & 92.07&	92.07&	91.37&	92.07&	\textbf{93.3}&	90.13&	92.6&	92.78&	91.19&	\textbf{93.66}&	91.72&	92.07&	92.25&	92.43&	91.9&	92.43\\
\hline
\textbf{Wine} & 95.5 & 88.75&	87.60&	93.80&	92.12&	88.17&	88.74&	92.69&	\textbf{96.64}&	89.85&	89.29&	92.13&	88.75&	91.56&	93.81&	89.30&	92.12\\
\hline
\end{tabular}

\caption{Performance of Logistic Regression on different outlier algorithms.}
\begin{tabular}{|c|c|c|c|c|c|c|c|c|c|c|c|c|c|c|c|c|c|c| } 
 \hline
 \textbf{Dataset} & \textbf{Original} & \textbf{ABOD} &	\textbf{COPOD} &	\textbf{CBLOF} &	\textbf{F.Bagging} & \textbf{HBOS} & \textbf{IForest} &	\textbf{KNN} & \textbf{Avg KNN} &	\textbf{LOF} &	\textbf{MCD} &	\textbf{OCSVM} &	\textbf{PCA} &	\textbf{LSCP} &	\textbf{COF} &	\textbf{SOD} &	\textbf{LODA}\\
 \hline
 \hline
\textbf{Annthyroid} & 95.1 & \textbf{95.25}&	\textbf{95.35} &	\textbf{95.25}&	95.07&	95.06&	\textbf{95.26}&	\textbf{95.35}&	\textbf{95.29}&	94.97&	95.03&	\textbf{95.14}&	\textbf{95.14}&	94.88&	95.00	&\textbf{95.50}&	\textbf{95.24}\\
\hline
\textbf{Arrhythmia} & 71.0  & 68.73	&66.74&	67.63&	66.74&	66.29&	67.4&	67.4&	67.18&	66.96&	68.07&	68.74&	67.4&	67.62&	68.51&	67.19&	66.74\\
\hline
\textbf{BreastW} & 95.7  &\textbf{ 95.98}	&95.55&	95.70 &	\textbf{96.13}&	95.70&	95.41&	\textbf{95.98}&	\textbf{95.84}&	95.42&	95.70&	\textbf{95.98}&	\textbf{95.84}&	95.13&	\textbf{95.98}&	95.41&	\textbf{95.98}\\
\hline
\textbf{Glass} & 59.2  & 57.75 &	57.75 &	\textbf{60.56} &	\textbf{60.09} &	\textbf{60.09} &	\textbf{62.44} &	58.22 &	59.15 &	59.15 &	59.15 &	\textbf{61.97} &	\textbf{60.56} &	57.28 &	\textbf{61.03} &	59.62 &	\textbf{60.09}\\
\hline
\textbf{Heart}& 78.3  & 76.40&	\textbf{80.52}&	\textbf{78.65}&	76.78&	74.91&	77.90&	\textbf{79.03} & 77.90&	78.28&\textbf{	80.52}&	78.28&	\textbf{78.65}&	73.03&\textbf{	80.52}	&75.66&	78.28\\
\hline
\textbf{Ionosphere} & 83.2  &\textbf{ 83.75}&	82.88&\textbf{	83.75}&	82.89&	\textbf{83.74}&	80.89&	81.17&	82.03&	83.18&	82.31&	80.89&	82.03&	83.18&	\textbf{83.46}&	\textbf{85.17}&	\textbf{83.74}\\
\hline
\textbf{Letter} & 76.0  & 75.99 &	75.78&	73.99&	75.78&	75.62&	74.92& \textbf{76.51}&	\textbf{76.48}&	75.58&	74.44&	75.76&	75.36&	75.83&	75.77&	\textbf{76.30}&	74.68\\
\hline
\textbf{Lympho} & 83.0  & \textbf{85.03 }&	82.31	&82.31&	80.95&	82.31&	78.23&	77.55&	82.31&	79.59&	82.99&	80.95&	77.55&	80.27&	78.23&	\textbf{83.67}&	82.31\\
\hline 
\textbf{Optdigits} & 96.5  &96.37&	\textbf{96.64}&	95.82&	96.1&	96.48&	96.42&	96.05&	95.91&	96.05&	96.35&	\textbf{96.76}&	96.42&	96.09&	95.78&	96.35&	96.41\\
\hline
\textbf{Pendigits}& 93.5  & \textbf{94.15} &	92.46&	90.51&	93.31&	\textbf{93.82}&	91.78&	\textbf{93.83}&	\textbf{93.58}&	\textbf{93.59}&	\textbf{94.93}&	\textbf{95.11}&	93.19&	93.25&	\textbf{93.89}&	\textbf{94.25}&	93.10 \\
\hline
\textbf{Satellite} & 79.2 & \textbf{79.46}&	78.79&	75.94	&78.55&	78.97&	78.59&	78.17&	78.35&	78.37&	78.46&	\textbf{79.29}&	79.04&	78.63&	78.99&	\textbf{79.38}&	78.87 \\
\hline
\textbf{Vertebral} & 83.6  & \textbf{83.86}&	\textbf{83.85}&	\textbf{83.85}&	\textbf{84.18}&\textbf{	83.86}&	\textbf{84.18}&	\textbf{83.85}&	\textbf{84.18}&	\textbf{84.18}&	\textbf{83.85}&	83.53&	82.57 &	82.57&	83.21&	\textbf{84.18}&	83.53\\
\hline
\textbf{WBC} & 94.3  & \textbf{94.54}&	94.01&	93.83&	94.01&	94.72	&\textbf{94.89}&	\textbf{94.36}&	\textbf{94.54}&	\textbf{94.54}&	\textbf{94.54}&	93.83&	93.66&	\textbf{94.89}&	\textbf{94.89}&	\textbf{94.54}&	\textbf{94.36}\\
\hline
\textbf{Wine} & 93.8  & 93.25&	\textbf{93.83}&	93.25&	92.13&	92.14&	92.15&	\textbf{94.38}&	\textbf{94.94}&	91.55&	92.70&	\textbf{93.81}&	\textbf{93.81}&	\textbf{94.93}&	92.68&	92.12&	\textbf{93.81}\\
\hline
\end{tabular}
% \end{table}
\end{sidewaystable}

\subsection{Discussion}
% vertebral - all classifier - all algo - improvement- less dim(only 6)
% breastw-  most of the models work good- pCA, COPOD, CBLOF, IForest- precision = 1 here.
% max samples 20000 in lympho- LR - KNN, SOD, RF- ABOD, HBOS, IF, PCA, LODA
% pendigits- RF- 100% 
% ABOD - on any classifier - dataset---- annthyroid, breastw, pendigits
% F. BAgging- not very good perf
% LODA -on DT and RF classifier - dataset---- lympho, pendigits, vertebral. 
% insights

Table 4 and 5 shows the Precision and Recall of 15 different algorithms on 20 different datasets. Values highlighted in bold shows that best algorithm for that particular dataset. By just looking the values, it is very difficult to predict which algorithm is best. Based on average precision and recall, we can comment that algorithm CBLOF with average precision and recall of 0.46 and 0.34 is doing better than other algorithms.

Table 6-8 shows the change in model performance after applying different outlier detection techniques. We have conducted this analysis on 14 different datasets. Values highlighted in bold indicates the performance value is greater than performance on original dataset. All the outlier techniques helping in the improvement of ML model at least for two datasets. Overall ABOD and OCVM techniques helps most in terms of model improvement across classifiers. If we look individual classifiers, then for RF classifier, ABOD, HBOS, OCSV, PCA, and LODA techniques performing better than other techniques. Similary for DT classifier, ABOD, CBLOF, HBOS, OCSVM, LSCP, and LODA giving better results as compare to other outlier detection algorithms. For LR classifier, ABOD, OCSVM, KNN, Avg KNN, SOD is performing better than other techniques. We can conclude that there is no single universally applicable or generic outlier detection approach. However, still this analysis can help data scientist to select outlier detection techniques. As a future scope of work, we would like to create new outlier detection technique which can capture the properties of different outlier detection techniques.

\section{Conclusion}
In this work, we have tried to provide details and compare a broad sample of current outlier techniques. We picked four different outlier categorization: Proximity, Linear Model, Ensemble, and Probabilistic. However, covering all the possible outlier techniques in single paper is very difficult, so we selected few techniques from above mentioned categorization. We have compared the different outlier detection techniques using precision and recall. Also, we have shown the change in model performance after removing the detected outliers by each of the outlier detection algorithm. We find that there is no single universally applicable or generic outlier detection approach.

\label{conc}
% This paper gives the comparison between different outlier detection techniques using standard global datasets from different domains. The paper contains results of two tasks, one is the calculation of precision score and recall score of outliers detected and other is to see the effect of outlier removal on accuracy of classification models. It is seen that the test accuracy of different classifiers used(Logistic Regression, Decision Trees and Random Forests) improve, when we remove the outliers. When the outliers are removed, the train data becomes less but still the accuracy on the unseen dataset is increasing. From this we can conclude that the model is misguided by the outliers. When outliers are removed, model better understands the underlying data distribution and predicts better on unseen test data. This also agrees with the fact that outliers have a disproportionate effect on result which can result in misleading interpretations.

\bibliographystyle{splncs04}
\bibliography{biblograph}

\end{document}